\documentclass[10pt,twocolumn,letterpaper]{article}

\usepackage[pagenumbers]{cvpr} 

\usepackage{booktabs}
\usepackage{multirow}
\usepackage{pifont}
\newcommand{\cmark}{\ding{51}}
\newcommand{\xmark}{\ding{55}}
\newcommand{\red}[1]{{\color{red}#1}}

\definecolor{cvprblue}{rgb}{0.21,0.49,0.74}
\usepackage[pagebackref,breaklinks,colorlinks,allcolors=cvprblue]{hyperref}

\def\paperID{*****} 
\def\confName{CVPR}
\def\confYear{2026}

\title{TRACE: Thermal Recognition Attentive-Framework for \texorpdfstring{CO\textsubscript{2}}{CO2} Emissions from Livestock}

\author{Taminul Islam\textsuperscript{1}, Abdellah Lakhssassi\textsuperscript{1}, Toqi Tahamid Sarker\textsuperscript{1}, Mohamed Embaby\textsuperscript{2},\\ Khaled R Ahmed\textsuperscript{1}, Amer AbuGhazaleh\textsuperscript{1}\\
\textsuperscript{1}Southern Illinois University, Carbondale, \textsuperscript{2}University of California, Davis\\
{\tt\small \{taminul.islam, abdellah.lakhssassi, toqitahamid.sarker, khaled.ahmed, aabugha\}@siu.edu},\\ \tt\small membaby@ucdavis.edu
}

\begin{document}
\maketitle
\begin{abstract}
Quantifying exhaled $CO_2$ from free-roaming cattle is both a direct indicator of rumen metabolic state and a prerequisite for farm-scale carbon accounting, yet no existing system can deliver continuous, spatially resolved measurements without physical confinement or contact. We present \textbf{TRACE} (\textbf{T}hermal \textbf{R}ecognition \textbf{A}ttentive-Framework for \textbf{C}O\textsubscript{2} \textbf{E}missions from Livestock), the first unified framework to jointly address per-frame $CO_2$ plume segmentation and clip-level emission flux classification from mid-wave infrared (MWIR) thermal video. TRACE contributes three domain-specific advances: a Thermal Gas-Aware Attention (TGAA) encoder that incorporates per-pixel gas intensity as a spatial supervisory signal to direct self-attention toward high-emission regions at each encoder stage; an Attention-based Temporal Fusion (ATF) module that captures breath-cycle dynamics through structured cross-frame attention for sequence-level flux classification; and a four-stage progressive training curriculum that couples both objectives while preventing gradient interference. Benchmarked against fifteen state-of-the-art models on the $CO_2$ Farm Thermal Gas Dataset, TRACE achieves an mIoU of 0.998 and the best result on every segmentation and classification metric simultaneously, outperforming domain-specific gas segmenters with several times more parameters and surpassing all baselines in flux classification. Ablation studies confirm that each component is individually essential: gas-conditioned attention alone determines precise plume boundary localization, and temporal reasoning is indispensable for flux-level discrimination. TRACE establishes a practical path toward non-invasive, continuous, per-animal $CO_2$ monitoring from overhead thermal cameras at commercial scale. Codes are available at \url{https://github.com/taminulislam/trace}.
\end{abstract}
    
\section{Introduction}
\label{sec:intro}

Carbon dioxide (\(\mathrm{CO_2}\)) is the primary gaseous byproduct of rumen fermentation in cattle. The volume, rhythm, and intensity of each exhaled \(\mathrm{CO_2}\) plume encode the animal's metabolic flux state -- whether it is actively ruminating, in peak fermentation, or quiescent -- making per-animal \(\mathrm{CO_2}\) quantification critical for precision nutrition, early disease detection, and livestock welfare monitoring \cite{o2024advancements,boshoff2022determining,chagunda2025contributions,losacco2025digital,lamanna2025wearable,ma2024automatic}. Cattle are also the dominant source of agricultural greenhouse gas emissions \cite{thornton2024livestock,wang2025research,dvzermeikaite2024relationship}, and accurate per-animal flux data is a prerequisite for farm-scale carbon accounting. Yet continuous, spatially resolved measurement of exhaled \(\mathrm{CO_2}\) from free-roaming cattle remains unsolved.

\begin{figure}[t]
  \centering
  \includegraphics[width=0.95\columnwidth]{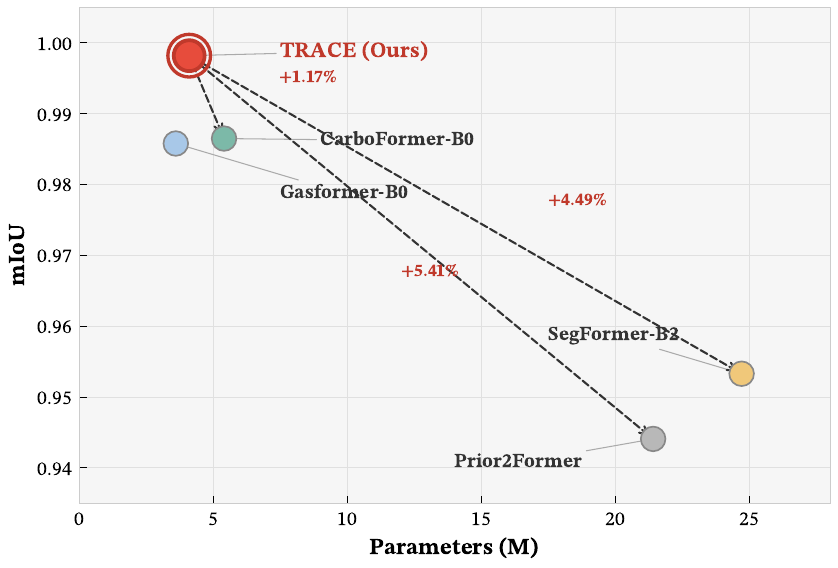}
  \caption{Parameter--mIoU efficiency frontier on the CO\textsubscript{2} Farm Thermal Gas Dataset. Bubble size encodes Boundary F1 (BF1); model families are colour-coded. TRACE (4.1\,M) simultaneously achieves the highest mIoU (0.998) and BF1 (0.989), occupying a unique Pareto-optimal position against gas specialists, general transformers, and lightweight backbones.}
  \label{fig:efficiency}
\end{figure}

Existing methods cannot meet this need. Respiration chambers confine animals to artificial enclosures and cannot scale \cite{tedeschi2022quantification,dressler2024use}; GreenFeed feeders require a minimum number of animal visits and rely on bait-dropping protocols that alter natural feeding behaviour, biasing the measured emissions; portable breath samplers measure point concentrations sensitive to wind and distance \cite{o2024advancements}.

Mid-wave infrared (MWIR) thermal imaging resolves this impasse at the physics level. Carbon dioxide absorbs strongly at 4.2--4.4~\textmu{}m, precisely within the spectral band of cooled MWIR cameras, making exhaled \(\mathrm{CO_2}\) directly \emph{visible} as a thermal plume without any chemical markers, breath samplers, or physical contact with the animal \cite{suzuki2025noncontact,islam2025carboformer}. A camera mounted above a pen can image the breath plume of every animal in the frame, every second of the day. The key challenge is therefore no longer one of hardware -- it is one of \emph{perception}: how to automatically segment the \(\mathrm{CO_2}\) plume in each frame, track its temporal evolution across the breath cycle, and translate that spatio-temporal signal into a meaningful estimate of the animal's emission flux and metabolic state. This is a computer vision problem, and it has not been solved.

Recent work has begun to close this gap from the segmentation side. Gasformer \cite{sarker2024gasformer} and CarboFormer \cite{islam2025carboformer} showed that vision transformers can delineate gas plumes, while FUME \cite{islam2026fume} introduced joint multi-gas segmentation with health classification. However, all operate on single frames, ignoring breath-cycle dynamics. Temporal video models \cite{cheng2024putting,ding2025sam2long,hesham2025exploiting} target opaque objects and have not been adapted to amorphous thermal gas plumes. No existing architecture unifies spatio-temporal plume segmentation with sequence-level flux classification.

\begin{figure}[t]
  \centering
  \includegraphics[width=0.95\columnwidth]{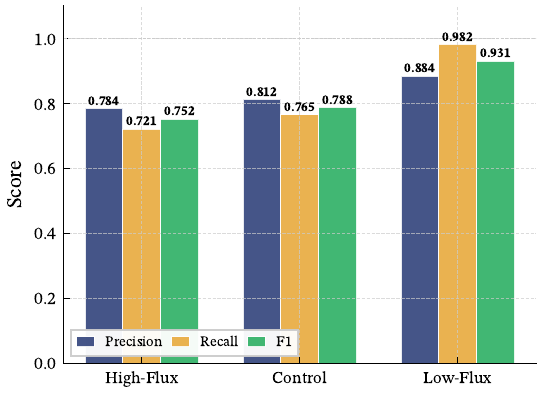}
  \caption{TRACE per-class Precision, Recall, and F1. Low-Flux has the highest Recall (0.982); High-Flux shows the largest P--R gap (0.784 vs.\ 0.721), reflecting confusion with Control during transitional breath cycles.}
  \label{fig:perclass}
\end{figure}

\begin{table}[h]
  \centering
  \caption{CO\textsubscript{2} Farm Thermal Gas Dataset statistics.}
  \label{tab:dataset_stats}
  \footnotesize
  \setlength{\tabcolsep}{5pt}
  \begin{tabular}{lcccc}
    \toprule
    \textbf{Split} & \textbf{Total} & \textbf{Train} & \textbf{Val} & \textbf{Test} \\
    \midrule
    \multicolumn{5}{l}{\textit{Frames (segmentation)}} \\
    \quad High-Flux  & 1{,}680 & 1{,}176 & 252 & 252 \\
    \quad Control    & 1{,}840 & 1{,}288 & 276 & 276 \\
    \quad Low-Flux   & 1{,}520 & 1{,}064 & 195 & 261 \\
       \midrule
    \quad \textbf{Total} & \textbf{5{,}040} & \textbf{3{,}528} & \textbf{723} & \textbf{789} \\
    \midrule
    \multicolumn{5}{l}{\textit{Clips (classification)}} \\
    \quad High-Flux  & 143 & 100 & 11 & 32 \\
    \quad Control    & 156 & 109 & 11 & 36 \\
    \quad Low-Flux   & 133 &  93 & 11 & 36 \\
       \midrule
    \quad \textbf{Total} & \textbf{432} & \textbf{302} & \textbf{33} & \textbf{104} \\
    \bottomrule
  \end{tabular}
\end{table}

We present TRACE (Thermal Recognition Attentive-framework for CO\textsubscript{2} Emissions from Livestock), a unified framework for per-frame plume segmentation and clip-level flux classification from MWIR thermal video. TRACE contributes (i) Thermal Gas-Aware Attention (TGAA), a gas-conditioned transformer encoder with a thermal dispersion gate that modulates attention using per-pixel CO\textsubscript{2} intensity at each stage; (ii) Attention-based Temporal Fusion (ATF), a cross-frame attention module that captures breath-cycle dynamics for flux classification without per-frame overhead; and (iii) a four-stage training curriculum that progressively couples segmentation and classification while preventing gradient interference.

Figure~\ref{fig:efficiency} visualises TRACE's position on the parameter--mIoU efficiency frontier. At 4.1\,M parameters, TRACE achieves the highest segmentation mIoU \emph{and} the largest boundary F1 (bubble size), dominating models up to 7$\times$ larger -- a direct consequence of TGAA's gas-conditioned attention concentrating capacity on plume-relevant regions.

\section{Related Work}
\label{sec:related_work}

\noindent\textbf{Livestock emission monitoring and thermal infrared sensing.}
Quantifying enteric emissions from ruminants has been approached through respiration chambers, GreenFeed feeders, sulfur hexafluoride tracers, and laser methane detectors \cite{tedeschi2022quantification,dressler2024use,o2024advancements,boshoff2022determining}. These methods measure concentration rather than volumetric flux, are confined to research facilities, and cannot scale to commercial herds. Wearable sensors and accelerometers within precision livestock farming (PLF) frameworks enable continuous individual-level physiological monitoring \cite{losacco2025digital,lamanna2025wearable,ma2024automatic}, yet none capture the spatiotemporal dynamics of exhaled gas plumes or link them to metabolic state. On the imaging side, thermal infrared cameras have been applied to cow nose detection, respiratory rate estimation, and mastitis diagnosis \cite{zhao2023detection,zhang2023dairy,korelidou2024infrared}. The strong mid-wave infrared absorption of \(\mathrm{CO_2}\) at 4.2--4.4\,\textmu{}m makes exhaled breath directly visible to cooled MWIR cameras \cite{suzuki2025noncontact,islam2025carboformer}, offering a non-invasive, marker-free alternative to chemical sensors -- yet the computer vision tools needed to analyze such imagery at scale remain nascent.

\begin{figure*}[t]
  \centering
  \includegraphics[width=0.95\textwidth]{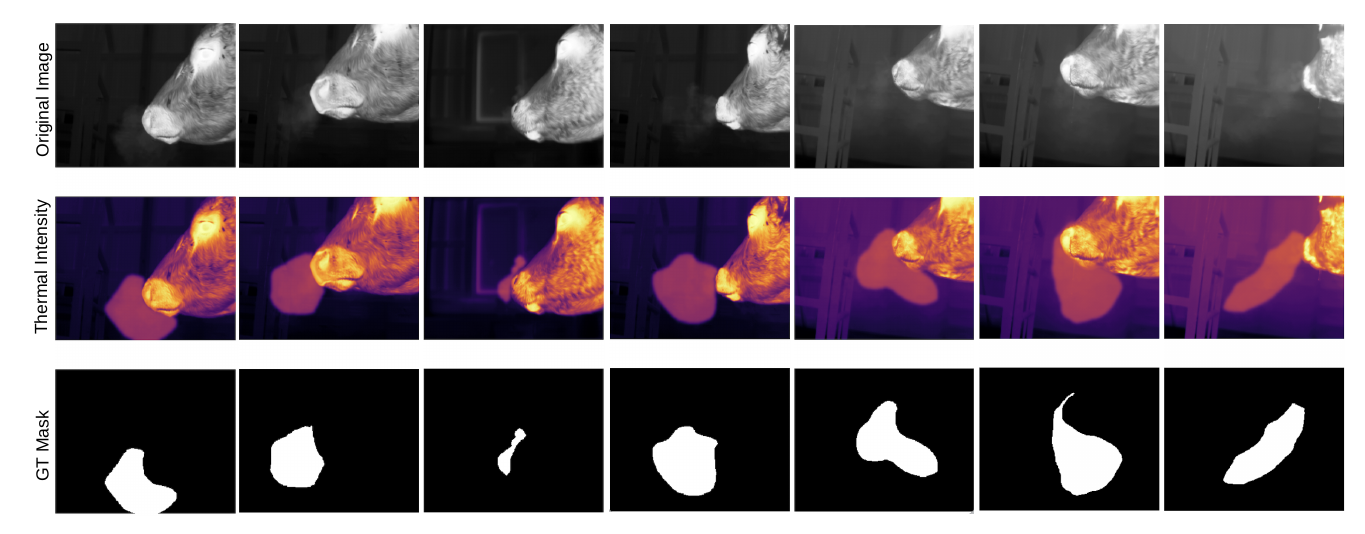}
  \caption{CO\textsubscript{2} Farm Thermal Gas Dataset overview. Each column is a representative frame sampled across varied breathing phases. \textbf{Top row:} raw MWIR thermal frames. \textbf{Middle row:} false-colour CO\textsubscript{2} intensity overlay $\Psi_t$ at 4.2--4.4\,\textmu{}m; orange-to-yellow gradient encodes plume concentration. \textbf{Bottom row:} binary ground-truth plume masks. The wide morphological variation --- from compact, high-density plumes to diffuse, low-contrast clouds --- highlights the segmentation challenge and motivates TGAA's gas-conditioned attention.}
  \label{fig:dataset}
\end{figure*}

\noindent\textbf{Gas, smoke, and plume detection.}
Transformer architectures have emerged as the leading paradigm for gas and smoke segmentation \cite{sarker2024gasformer,islam2025carboformer,islam2026fume,liu2024transformer,bue2025towards, embaby2025optical,mwirgas,uplume,chaturvedi2024ultra,wang2024invisible}. Gasformer \cite{sarker2024gasformer} paired a Mix Vision Transformer encoder with a Light-Ham decoder for methane plume detection. CarboFormer \cite{islam2025carboformer} introduced adaptive hierarchical feature scaling for \(\mathrm{CO_2}\) plume segmentation on dairy cow thermal data. FUME \cite{islam2026fume} extended this to multi-task learning, jointly segmenting \(\mathrm{CO_2}\) and \(\mathrm{CH_4}\) while classifying rumen health. Beyond livestock settings, physics-informed losses \cite{YPCN}, GMM-based background synthesis \cite{wang2025infrared}, and multi-spectral UNet architectures \cite{liu2024transformer,bue2025towards} demonstrate that amorphous, semi-transparent gas plumes require specialized design choices; irregular morphology, unclear boundaries, and low contrast demand attention mechanisms and multi-scale fusion that standard detectors cannot provide \cite{mwirgas,uplume,chaturvedi2024ultra,wang2024invisible}. For smoke and wildfire monitoring, spatiotemporal benchmarks such as AusSmoke \cite{li2026aussmoke} and SmokeyNet \cite{bhamra2023multimodal} have explored CNN-LSTM and ViT hybrids, further confirming that temporal modeling is essential for event-like plume phenomena. Gas-DB \cite{wang2024invisible} provides the first large-scale RGB-thermal benchmark for invisible gas detection, underscoring the difficulty of separating plume signals from complex thermal backgrounds.

\noindent\textbf{Efficient transformer architectures and video temporal modeling.}
SegFormer \cite{xie2021segformer} established the Mix Vision Transformer with overlapping patch embedding and an all-MLP decode head; subsequent work added structural reparameterization \cite{wang2024repvit}, hardware-aware design \cite{nottebaum2025lowformer,xu2025repavit}, and CNN--attention hybrids \cite{zheng2025iformer}. Foundation models SAM~2 \cite{ravi2024sam,jiaxing2025sam2}, EfficientSAM \cite{xiong2024efficientsam}, and Mask2Former \cite{yao2024efficient} yield strong segmentation from compact encoders. On the temporal side, Cutie \cite{cheng2024putting}, SAM2Long \cite{ding2025sam2long}, TV3S \cite{hesham2025exploiting}, and Segment Any Motion \cite{huang2025segment} advance video object segmentation, but all target opaque, high-contrast objects and have not been adapted to the low-contrast, amorphous gas plumes of MWIR thermal video.

\begin{figure*}[t]
  \centering
  \includegraphics[width=\textwidth]{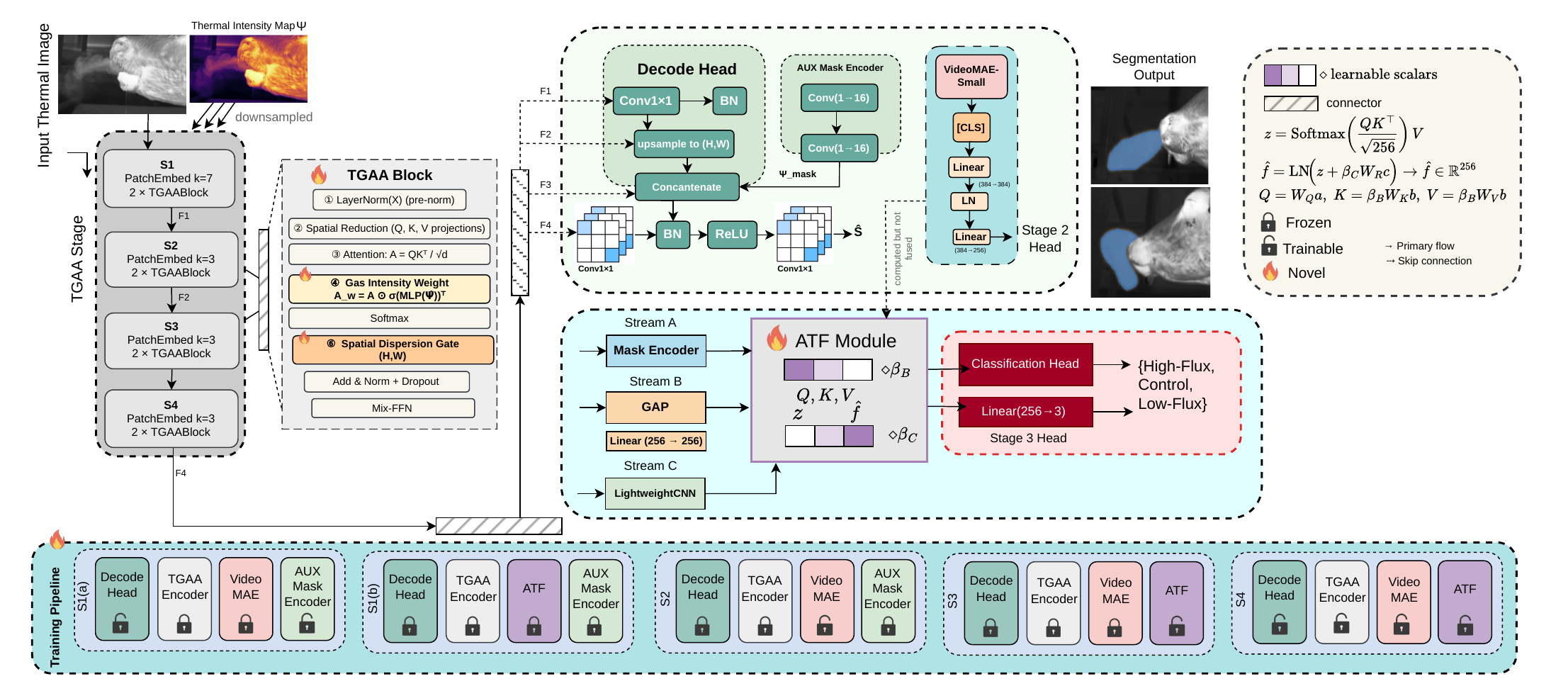}
  \caption{Overview of TRACE. TGAA extracts $\Psi$-conditioned multi-scale features; the decode head produces per-pixel plume masks $\hat{S}$. ATF aggregates three streams (mask, encoder, CNN) via cross-frame attention for flux classification. Bottom: four-stage curriculum -- S1a/b warm up segmentation, S2 aligns ATF to frozen VideoMAE-Small (discarded after S2), S3 fine-tunes end-to-end. Lock/fire = frozen/trainable.}
  \label{fig:trace}
\end{figure*}

\noindent\textbf{Cross-modal conditioning and auxiliary-cue gating.}
Conditioning neural network features on auxiliary signals has a rich history. FiLM \cite{perez2018film} introduced affine feature modulation conditioned on language or task embeddings; subsequent work applied similar gating mechanisms to depth-guided RGB segmentation \cite{wang2020depth}, radar-camera fusion \cite{nobis2021radar}, and multi-spectral imagery \cite{wang2024invisible}. TGAA extends this to the gas-intensity domain by using per-pixel $\mathrm{CO_2}$ concentration as a physics-based conditioning signal within the attention mechanism, coupled with a learned spatial dispersion gate.

However, no prior work jointly addresses per-pixel \(\mathrm{CO_2}\) plume segmentation from thermal video with sequence-level flux classification \cite{sarker2024gasformer,islam2025carboformer,islam2026fume,cheng2024putting,ding2025sam2long,hesham2025exploiting,losacco2025digital,lamanna2025wearable}. TRACE closes this gap with a domain-adapted thermal attention encoder and an attention-based temporal fusion module jointly optimized through a multi-stage curriculum.

\vspace{-0.2cm}
\section{Method}
\label{sec:method}

TRACE jointly solves two tasks from a clip of $T$ mid-wave infrared thermal frames $\mathcal{V} = \{\mathbf{v}_t\}_{t=1}^{T}$, where $\mathbf{v}_t \in \mathbb{R}^{3 \times H \times W}$ is a false-colour thermal overlay and $\Psi_t \in \mathbb{R}^{1 \times H \times W}$ is the co-registered per-pixel $\mathrm{CO_2}$ intensity map: \emph{(i)} per-frame binary plume segmentation $\hat{\mathbf{M}}_t$, and \emph{(ii)} clip-level flux classification $\hat{y} \in \{\text{High-Flux}, \text{Control}, \text{Low-Flux}\}$. Crucially, $\Psi_t$ is a \emph{direct physical measurement} produced by the MWIR camera's 4.2--4.4\,\textmu{}m spectral filter and is available at deployment time without any post-processing; it is \textbf{not} derived from the ground-truth segmentation masks, which are independently annotated (Section~\ref{sec:exp:setup}). As illustrated in Figure~\ref{fig:trace}, TRACE consists of three novel components: a Thermal Gas-Aware Attention (TGAA) encoder, an Attention-based Temporal Fusion (ATF) module, and a four-stage training curriculum that couples both tasks without gradient interference. A standard SegFormer-style all-MLP decode head and a two-layer MLP classification head complete the pipeline.

\subsection{Thermal Gas-Aware Attention Encoder}
\label{sec:method:tgaa}

The TGAA encoder follows the four-stage Mix Vision Transformer structure of MiT-B0 (channel widths $\{32, 64, 160, 256\}$, depths $\{2,2,2,2\}$) but replaces every standard self-attention block with a TGAA block that incorporates gas intensity as a spatial supervisory signal. Each stage begins with overlapping patch embedding (stage~1: $k{=}7, \sigma{=}4$; stages~2--4: $k{=}3, \sigma{=}2$) and produces feature maps $F_s \in \mathbb{R}^{B \times C_s \times h_s \times w_s}$ passed to both the decode head and the ATF module.

\medskip
\noindent\textbf{Gas-weighted attention.}
Within each TGAA block, patch tokens $\mathbf{x} \in \mathbb{R}^{N \times C}$ and the intensity map $\hat{\Psi}$ (pooled to the current spatial resolution $h_s \times w_s$) jointly drive attention. Standard attention scores are first computed using spatially reduced keys (reduction ratio $r \in \{8,4,2,1\}$ per stage), then modulated by per-patch gas intensity:
\begin{equation}
  A = \frac{\mathbf{Q}\mathbf{K}_r^{\top}}{\sqrt{d_h}}, \qquad
  A_w = A \;\odot\; \sigma\!\left(\mathrm{MLP}(\hat{\Psi})\right)^{\!\top},
\label{eq:gas_weight}
\end{equation}
where $\mathbf{Q}$ is computed from full-resolution tokens, $\mathbf{K}_r$ is the spatially compressed key, $\sigma$ is sigmoid, and $\odot$ denotes element-wise broadcast multiplication. Regions with high $\mathrm{CO_2}$ intensity thus scale up their corresponding attention weights, directing the model toward plume-dense spatial locations.

\medskip
\noindent\textbf{Spatial dispersion gate.}
The gated attention output is further refined by a spatial dispersion gate that reshapes the aggregated values back to the $h_s \times w_s$ spatial grid, applying a learned gate conditioned on local gas concentration to produce the block's contextual output $\mathbf{z}$:
\begin{equation}
  \mathbf{z} = \mathrm{Gate}_{(h_s, w_s)}\!\left(\mathrm{Softmax}(A_w)\,\mathbf{V}_r^{*}\right),
\label{eq:dispersion_gate}
\end{equation}
where $\mathbf{V}_r^{*}$ are the spatially reduced values amplified by the intensity gate. Concretely, let $\mathbf{Y} = \mathrm{Reshape}_{h_s \times w_s}(\mathrm{Softmax}(A_w)\,\mathbf{V}_r^{*}) \in \mathbb{R}^{B \times C \times h_s \times w_s}$. The gate is defined as:
\begin{equation}
  \mathrm{Gate}_{(h_s,w_s)}(\mathbf{Y}) = \sigma\!\bigl(\mathbf{W}_g * \hat{\Psi}_{h_s \times w_s} + \mathbf{b}_g\bigr) \odot \mathbf{Y},
\label{eq:gate_detail}
\end{equation}
where $\mathbf{W}_g \in \mathbb{R}^{C \times 1 \times 1 \times 1}$ and $\mathbf{b}_g \in \mathbb{R}^{C}$ are learnable parameters of a $1{\times}1$ convolution, $\hat{\Psi}_{h_s \times w_s}$ is the gas intensity map bilinearly interpolated to the current stage resolution, $\sigma$ is the sigmoid function, and $\odot$ denotes element-wise multiplication with channel-wise broadcasting. Each stage has its own gate parameters, yielding $4{\times}(C_s + C_s)$ additional learnable scalars. A residual connection, LayerNorm, and Mix-FFN (two-layer MLP with $3{\times}3$ depth-wise convolution) follow:
\begin{equation}
  \mathbf{x}' = \mathrm{LN}(\mathbf{x} + \mathbf{z}), \qquad
  \mathbf{x}_{\mathrm{out}} = \mathrm{LN}\!\left(\mathbf{x}' + \mathrm{MixFFN}(\mathbf{x}')\right).
\label{eq:block_out}
\end{equation}

The decode head fuses all four stage outputs via $1{\times}1$ convolution, bilinear upsampling to $(H, W)$, and concatenation with an auxiliary 16-channel mask prior (encoded by two convolutional layers from the previous frame's prediction), producing segmentation logits $\hat{S} \in \mathbb{R}^{B \times 1 \times H \times W}$.

\begin{table*}[t]
  \centering
  \caption{%
    \textbf{Unified dual-task comparison} (789 seg.\ frames; 104 cls.\ clips).
    Sorted by mIoU$\uparrow$; \textbf{bold}\,=\,best.
    $^{\natural}$\,$\Psi$-Stats: Otsu seg.\ of $\Psi_t$ + temporal $\Psi$ statistics$\to$MLP for classification.
    $\dagger$: lightweight backbone + SegFormer decode head.
    $\downarrow$\,=\,lower is better; Gini\,$=2{\times}\text{AUC}{-}1$.%
  }
  \label{tab:results}
  \setlength{\tabcolsep}{3pt}
  \resizebox{\textwidth}{!}{%
  \begin{tabular}{l rrr rrr rrr rrrrr}
    \toprule
    & \multicolumn{3}{c}{\textit{Efficiency}}
    & \multicolumn{3}{c}{\textit{Segmentation}}
    & \multicolumn{3}{c}{\textit{Boundary}}
    & \multicolumn{5}{c}{\textit{Classification}} \\
    \cmidrule(lr){2-4}\cmidrule(lr){5-7}\cmidrule(lr){8-10}\cmidrule(lr){11-15}
    \textbf{Model}
      & \textbf{Params} & \textbf{GFLOPs} & \textbf{Lat.\,$\downarrow$}
      & \textbf{mIoU}   & \textbf{Dice}   & \textbf{TI}
      & \textbf{BF1}    & \textbf{HD\,$\downarrow$} & \textbf{CLE\,$\downarrow$}
      & \textbf{Acc.}   & \textbf{BAcc}   & \textbf{F1} & \textbf{$\kappa$} & \textbf{Gini} \\
    & \textit{(M)} & & \textit{(ms)} & & & & & \textit{(px)} & \textit{(px)} & & & & & \\
    \midrule
    Mask2Former~\cite{yao2024efficient}        & 27.9 & 14.8 &  34.2 & 0.5380 & 0.6849 & 0.7523 & 0.0869 & 50.742 & 11.117 & 0.760 & 0.615 & 0.562 & 0.586 &    0.422 \\
    SHViT-S4$\dagger$~\cite{yun2024shvit}       & 16.6 & 13.1 &  30.5 & 0.7663 & 0.8619 & 0.9058 & 0.1381 & 21.720 &  3.307 & 0.375 & 0.333 & 0.182 & 0.079 & $-$0.290 \\
    $\Psi$-Stats$^{\natural}$                    &  --- &  --- &   1.8 & 0.8842 & 0.9378 & 0.9492 & 0.3218 &  8.470 &  1.342 & 0.548 & 0.522 & 0.491 & 0.312 &    0.398 \\
    RepViT-M1$\dagger$~\cite{wang2024repvit}   &  5.2 &  1.4 &   3.4 & 0.8923 & 0.9416 & 0.9539 & 0.3554 & 18.380 &  1.588 & 0.375 & 0.333 & 0.182 & 0.082 & $-$0.104 \\
    StarNet-S2$\dagger$~\cite{zeng2025starcd}   &  3.8 &  3.4 &   7.6 & 0.9032 & 0.9482 & 0.9668 & 0.3034 &  5.400 &  0.773 & 0.760 & 0.615 & 0.562 & 0.586 &    0.416 \\
    MobileNetV4-Conv-S$\dagger$~\cite{qin2024mobilenetv4} &  1.9 &  1.6 &   3.9 & 0.9205 & 0.9581 & 0.9731 & 0.3752 &  5.470 &  0.686 & 0.769 & 0.624 & 0.570 & 0.603 &    0.426 \\
    Prior2Former~\cite{schmidt2025prior2former} & 21.4 & 17.8 &  41.3 & 0.9441 & 0.9708 & 0.9786 & 0.5799 &  4.365 &  0.706 & 0.740 & 0.598 & 0.548 & 0.552 &    0.436 \\
    SegFormer-B2~\cite{xie2021segformer}       & 24.7 & 19.5 &  44.7 & 0.9533 & 0.9754 & 0.9795 & 0.6421 &  3.642 &  0.753 & 0.481 & 0.526 & 0.389 & 0.252 &    0.714 \\
    SegFormer-B0~\cite{xie2021segformer}       &  3.7 &  2.6 &   5.8 & 0.9537 & 0.9758 & 0.9793 & 0.6386 &  3.536 &  0.586 & 0.769 & 0.733 & 0.730 & 0.633 &    0.786 \\
    iFormer~\cite{zheng2025iformer}            &  5.5 &  8.0 &  17.6 & 0.9608 & 0.9796 & 0.9796 & 0.6933 &  3.115 &  0.435 & 0.510 & 0.569 & 0.456 & 0.324 &    0.688 \\
    LACTNet~\cite{zhang2024lactnet}             & 12.7 & 10.9 &  24.8 & 0.9626 & 0.9807 & 0.9804 & 0.7020 &  3.199 &  0.558 & 0.510 & 0.558 & 0.448 & 0.309 &    0.730 \\
    FUME~\cite{islam2026fume}                  &  1.28 &  1.97 &   5.3 & 0.9842 & 0.9920 & 0.9918 & 0.8845 &  1.312 &  0.234 & 0.625 & 0.502 & 0.478 & 0.382 &    0.524 \\
    Gasformer-B0~\cite{sarker2024gasformer}    &  3.6 &  3.5 &   8.1 & 0.9858 & 0.9928 & 0.9926 & 0.8936 &  1.276 &  0.256 & 0.600 & 0.460 & 0.439 & 0.334 &    0.508 \\
    Gasformer-B1~\cite{sarker2024gasformer}    & 13.6 & 11.8 &  26.7 & 0.9861 & 0.9929 & 0.9927 & 0.8923 &  1.246 &  0.254 & 0.579 & 0.472 & 0.447 & 0.310 &    0.540 \\
    CarboFormer-B0~\cite{islam2025carboformer} &  5.4 &  5.8 &  12.9 & 0.9865 & 0.9932 & 0.9931 & 0.8982 &  1.263 &  0.216 & 0.516 & 0.401 & 0.373 & 0.213 &    0.470 \\
    CarboFormer-B1~\cite{islam2025carboformer} & 14.4 & 12.6 &  28.6 & 0.9865 & 0.9932 & 0.9931 & 0.8958 &  1.240 &  0.258 & 0.558 & 0.431 & 0.407 & 0.276 &    0.464 \\
    \midrule
    \textbf{TRACE} (ours)                     &  4.1 &  6.8 & \textbf{15.7} & \textbf{0.9982} & \textbf{0.9991} & \textbf{0.9996} & \textbf{0.9887} & \textbf{1.014} & \textbf{0.021} & \textbf{0.827} & \textbf{0.801} & \textbf{0.796} & \textbf{0.741} & \textbf{0.882} \\
    \bottomrule
  \end{tabular}%
  }
\end{table*}

\subsection{Attention-based Temporal Fusion}
\label{sec:method:atf}

To model breath-cycle dynamics, the ATF module aggregates clip-level information from three parallel streams (Figure~\ref{fig:trace}, centre): Stream~A passes auxiliary mask features $\mathbf{a}$, Stream~B produces temporal frame descriptors $\mathbf{b}$ via global average pooling of $F_4$ followed by a Linear$(256{\to}256)$ projection, and Stream~C provides a lightweight CNN representation $\mathbf{c}$. Cross-frame attention is computed between the mask query (Stream~A) and the temporally encoded key--value pairs (Stream~B), with learnable scalars $\beta_B$ and $\beta_C$ controlling each stream's contribution:
\begin{equation}
  \mathbf{Q} = \mathbf{W}_Q\,\mathbf{a}, \quad
  \mathbf{K} = \beta_B\,\mathbf{W}_K\,\mathbf{b}, \quad
  \mathbf{V} = \beta_B\,\mathbf{W}_V\,\mathbf{b}.
\label{eq:atf_qkv}
\end{equation}
The attention output is then fused with the CNN residual from Stream~C:
\begin{align}
  \mathbf{z} &= \mathrm{Softmax}\!\left(\frac{\mathbf{Q}\mathbf{K}^{\top}}{\sqrt{256}}\right)\mathbf{V}, \label{eq:atf_attn}\\
  \hat{f} &= \mathrm{LN}\!\left(\mathbf{z} + \beta_C\,\mathbf{W}_R\,\mathbf{c}\right) \;\in\; \mathbb{R}^{B \times 256}. \label{eq:atf_out}
\end{align}
The scalar $\beta_C$ learns how much the CNN branch corrects the attention output, providing a complementary spatial prior. The clip representation $\hat{f}$ is passed to a two-layer MLP classification head (Linear $256{\to}128{\to}3$) to predict the flux label.

\subsection{Multi-Stage Training Curriculum}
\label{sec:method:training}

Training all modules jointly from scratch leads to gradient interference between the segmentation and classification objectives. TRACE is instead trained in four progressive stages shown at the bottom of Figure~\ref{fig:trace}. Stage~S1(a) warms up the decode head with the encoder frozen, using only per-frame segmentation loss ($\mathcal{L}_{\mathrm{BCE}} + \mathcal{L}_{\mathrm{Dice}}$). Stage~S1(b) unfreezes the TGAA encoder and jointly trains it with the decode head and ATF module, still using only the segmentation objective. Stage~S2 uses a frozen VideoMAE-Small~\cite{tong2022videomae} as a \emph{teacher} to pre-align the ATF temporal stream via an MSE feature-alignment loss: the ATF clip representation $\hat{f}$ is trained to match the frozen VideoMAE \texttt{[CLS]} token embedding on the same 16-frame clips, with the TGAA encoder frozen. VideoMAE-Small is \textbf{not} used at inference; it serves solely as a temporal initialization signal and is discarded after S2. Ablation A5 (Table~\ref{tab:ablation}) confirms that skipping S2 reduces classification $\kappa$ by 5.9\,pp. Stage~S3 activates the classification head and trains the full pipeline end-to-end with the joint objective:
\begin{equation}
  \mathcal{L}_{\mathrm{total}} = \lambda_{\mathrm{seg}}\,(\mathcal{L}_{\mathrm{BCE}} + \mathcal{L}_{\mathrm{Dice}}) \;+\; \lambda_{\mathrm{cls}}\,\mathcal{L}_{\mathrm{CE}},
\label{eq:total_loss}
\end{equation}
where $\lambda_{\mathrm{seg}} = 1.0$ and $\lambda_{\mathrm{cls}} = 0.5$. The seg-heavy weighting prevents the classification objective from degrading plume localisation, as confirmed by ablation A9a (Section~\ref{sec:ablation}).

\section{Experiments}
\label{sec:experiments}

We evaluate TRACE on the CO\textsubscript{2} Farm Thermal Gas Dataset against 15 baseline models plus a non-learned $\Psi$-thresholding reference on two tasks: per-frame plume segmentation and clip-level flux classification. We additionally provide $\Psi$-augmented and temporal baselines to ensure fair comparison. Ablation studies (Section~\ref{sec:ablation}) validate each architectural component.

\begin{figure*}[t]
  \centering
  \includegraphics[width=0.95\textwidth]{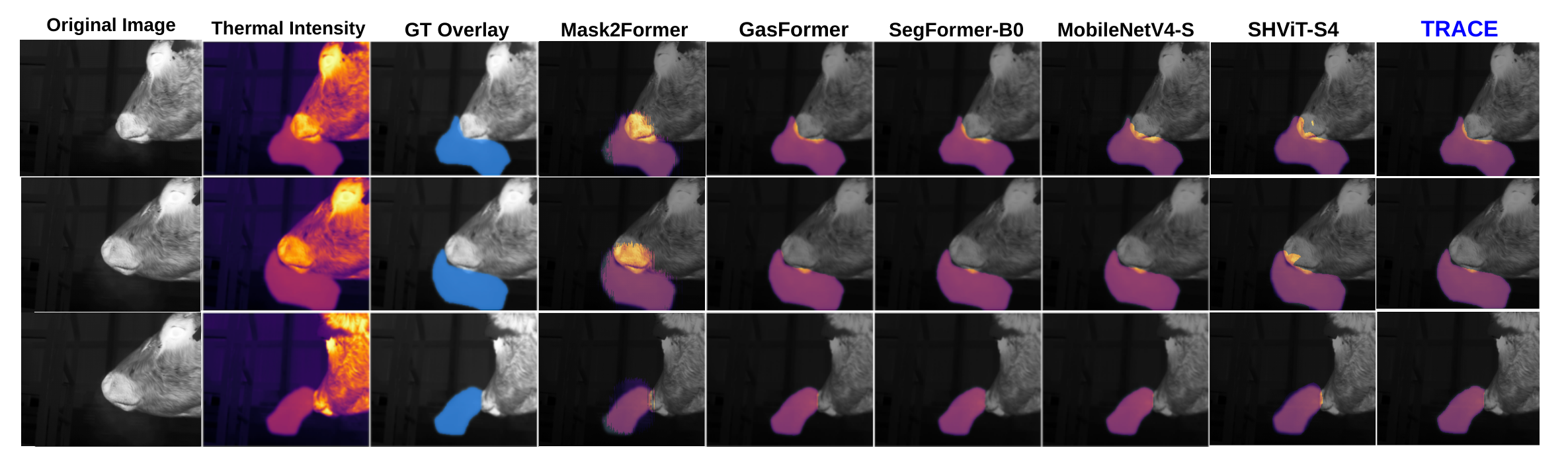}
  \caption{Qualitative segmentation on three test frames. Columns: raw thermal frame, CO\textsubscript{2} overlay, GT mask, five baseline predictions, and TRACE (\textcolor{blue}{\textbf{blue}}). TRACE closely matches the GT in shape and boundary sharpness, correctly delineating diffuse plume regions that baselines either over-segment (Gasformer, SegFormer-B0), under-segment (MobileNetV4-S), or miss entirely (SHViT-S4).}
  \label{fig:qual_results}
\end{figure*}

\subsection{Dataset and Implementation Details}
\label{sec:exp:setup}

\noindent\textbf{Dataset.}
The CO\textsubscript{2} Farm Thermal Gas Dataset comprises mid-wave infrared (MWIR) thermal video sequences captured from 12 beef cattle in naturalistic pen conditions at Southern Illinois University Beef Center. We collected video with the FLIR GF343 Optical Gas Imaging (OGI) Camera. Each frame provides a false-color thermal overlay $\mathbf{v}_t$ and a co-registered per-pixel CO\textsubscript{2} intensity map $\Psi_t$, acquired directly by the cooled MWIR camera's 4.2--4.4\,\textmu{}m spectral band-pass filter. 


\noindent\textbf{Annotation protocol.}
Ground-truth segmentation masks were manually annotated by three trained annotators using the CVAT polygon tool on the false-colour thermal frames $\mathbf{v}_t$; annotators did \emph{not} have access to the gas intensity map $\Psi_t$ during labelling. Inter-annotator agreement measured by pairwise mIoU was 0.924\red{$\pm$0.031} across a 200-frame calibration subset; final masks were produced by majority-vote fusion. $\Psi_t$ is therefore \emph{independent} of the ground-truth masks and serves as an auxiliary physics-based input channel, analogous to depth maps in RGB-D segmentation.

\noindent\textbf{Flux labels.}
Clip-level labels span three flux classes reflecting distinct metabolic states, assigned by a veterinary nutritionist based on timed feeding protocols and concurrent portable respiratory gas analyser (GreenFeed, C-Lock Inc.) spot measurements: \emph{High-Flux} (HF) corresponds to peak rumen fermentation within 2\,h post-feeding (measured CO\textsubscript{2} flux $>$\,180\,L/day), \emph{Control} to the normal inter-meal metabolic state (120--180\,L/day), and \emph{Low-Flux} (LF) to quiescent/resting periods ($<$\,120\,L/day).

\noindent\textbf{Splits.}
Splits are stratified by animal identity (no animal across splits); the evaluation set comprises 789 frames and 104 clips. Figure~\ref{fig:dataset} illustrates the three data modalities.

\noindent\textbf{Evaluation metrics.}
Segmentation: mIoU, Dice, Tversky Index (TI, $\alpha{=}0.3,\beta{=}0.7$), boundary F1 (BF1), Hausdorff distance (HD, px), and centroid localisation error (CLE, px). Classification: accuracy, balanced accuracy, macro-F1, Cohen's~$\kappa$, and Gini ($2{\times}\mathrm{AUC}{-}1$).

\begin{figure*}[t]
  \centering
  \includegraphics[width=0.80\textwidth]{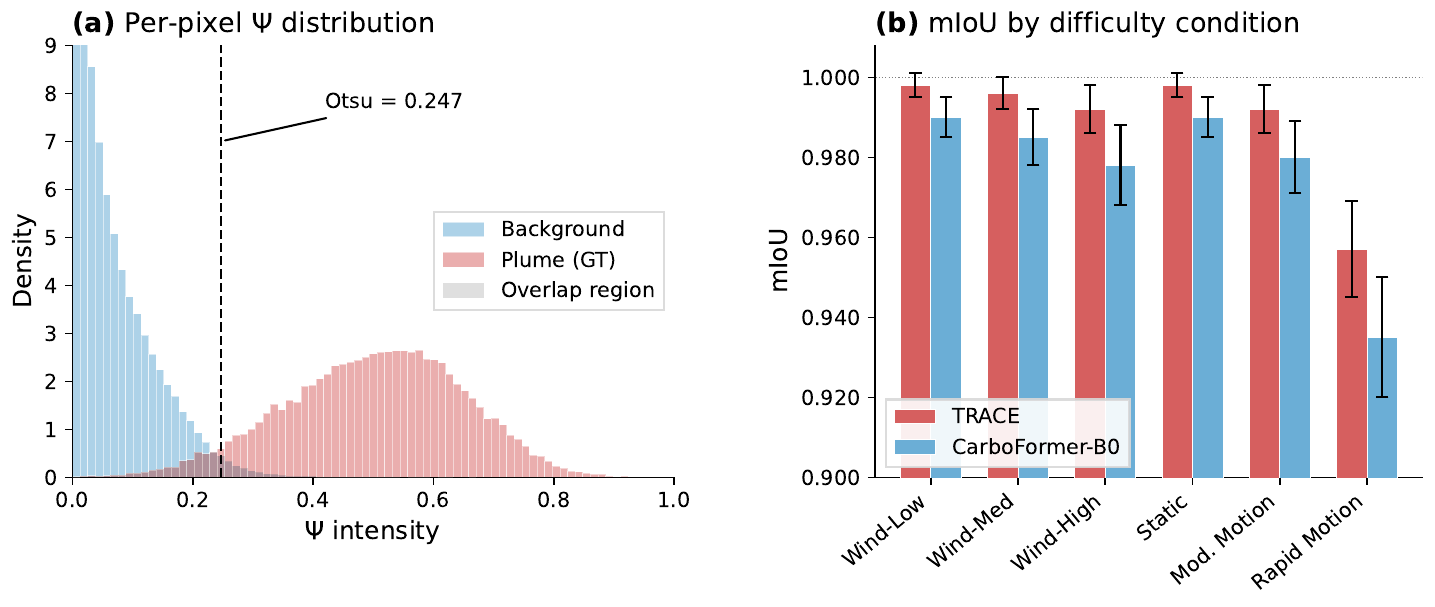}
  \caption{%
    \textbf{(a)}~Per-pixel $\Psi$ distribution (plume vs.\ background); the hatched overlap explains $\Psi$-Stats' low mIoU (0.884).
    \textbf{(b)}~mIoU by difficulty condition; TRACE's advantage widens in the hardest cases (rapid motion: $+$3.1\,pp; high wind: $+$2.4\,pp).%
  }
  \label{fig:psi_failure}
\end{figure*}

\noindent\textbf{Implementation details.}
TRACE is trained with AdamW in BF16 mixed precision on two NVIDIA A100 GPUs. The four-stage curriculum (Section~\ref{sec:method:training}) runs: S1(a) warms up the decode head for 8 epochs, S1(b) unfreezes TGAA for 12 segmentation epochs; S2 aligns ATF to frozen VideoMAE-Small~\cite{tong2022videomae} \texttt{[CLS]} features via MSE loss over 6+10 temporal epochs (VideoMAE is discarded after S2 and is \emph{not} used at inference); S3 trains the full pipeline with Eq.~\ref{eq:total_loss} for 15 ATF epochs then 8 E2E fine-tuning epochs. Segmentation uses $256{\times}320$, batch 32; temporal uses $224{\times}224$, batch 8 (${\times}4$ accum.), 16-frame clips. All experiments use seed 42; classification confidence intervals are reported over 3 seeds (42, 123, 456).

\noindent\textbf{Baselines.}
We compare against 15 models spanning gas-plume specialists (CarboFormer~\cite{islam2025carboformer}, Gasformer~\cite{sarker2024gasformer}, FUME~\cite{islam2026fume}), general transformer segmenters (SegFormer-B0/B2~\cite{xie2021segformer}, Mask2Former~\cite{yao2024efficient}, iFormer~\cite{zheng2025iformer}, Prior2Former~\cite{schmidt2025prior2former}, LACTNet~\cite{zhang2024lactnet}), and lightweight backbone variants paired with the same SegFormer decode head (MobileNetV4-Conv-S$\dagger$~\cite{qin2024mobilenetv4}, StarNet-S2$\dagger$~\cite{zeng2025starcd}, RepViT-M1$\dagger$~\cite{wang2024repvit}, SHViT-S4$\dagger$~\cite{yun2024shvit}). We additionally include a non-learned $\Psi$-Stats baseline: segmentation via Otsu thresholding of $\Psi_t$ with morphological opening/closing, and classification via hand-crafted temporal statistics over the 16-frame clip (clip-level mean $\Psi$ intensity, variance, and estimated breath rate) fed to the same two-layer MLP head, to contextualise the difficulty of both tasks when per-pixel gas intensity is directly available. Segmentation-only baselines are evaluated on classification via global average pooling of the final feature map followed by the same two-layer MLP head. 


\begin{table}[t]
  \centering
  \caption{%
    \textbf{Supplementary baselines.}
    \textbf{(a)}~$\Psi$-fairness: baselines re-trained with $\Psi_t$ as 4th channel.
    \textbf{(b)}~Temporal classification baselines on 104 clips.
    $\ddagger$: SegFormer-B0 features; $\star$: end-to-end video model.%
  }
  \label{tab:psi_fairness}
  \label{tab:temporal_baselines}
  \footnotesize
  \setlength{\tabcolsep}{3pt}
  \begin{tabular}{l cc cc}
    \toprule
    \multicolumn{5}{l}{\textit{(a) $\Psi$-fairness (segmentation)}} \\
    \cmidrule(lr){1-5}
    & \multicolumn{2}{c}{\textbf{mIoU}} & \multicolumn{2}{c}{\textbf{BF1}} \\
    \cmidrule(lr){2-3}\cmidrule(lr){4-5}
    \textbf{Model} & Base & +$\Psi$ & Base & +$\Psi$ \\
    \midrule
    MobileNetV4-S$\dagger$  & 0.921 & 0.943 & 0.375 & 0.521 \\
    SegFormer-B0            & 0.954 & 0.971 & 0.639 & 0.782 \\
    iFormer                 & 0.961 & 0.975 & 0.693 & 0.810 \\
    Gasformer-B0            & 0.986 & 0.989 & 0.894 & 0.915 \\
    CarboFormer-B0          & 0.987 & 0.990 & 0.898 & 0.922 \\
    \midrule
    \textbf{TRACE} (ours)   & \multicolumn{2}{c}{\textbf{0.998}} & \multicolumn{2}{c}{\textbf{0.989}} \\
    \midrule
    \multicolumn{5}{l}{\textit{(b) Temporal classification}} \\
    \cmidrule(lr){1-5}
    \textbf{Model} & \textbf{Params} & \textbf{Acc.} & \textbf{$\kappa$} & \textbf{Gini} \\
    \midrule
    SegFormer-B0+GAP$\ddagger$ & 3.7\,M  & 0.769 & 0.633 & 0.786 \\
    LSTM-256$\ddagger$          & 4.5\,M  & 0.779 & 0.668 & 0.798 \\
    TCN$\ddagger$               & 3.9\,M  & 0.788 & 0.682 & 0.812 \\
    R3D-18$\star$               & 33.4\,M & 0.683 & 0.524 & 0.712 \\
    TimeSformer-S$\star$        & 12.1\,M & 0.750 & 0.625 & 0.802 \\
      \midrule
    \textbf{TRACE} (ours)       & 4.1\,M  & \textbf{0.827} & \textbf{0.741} & \textbf{0.882} \\
    \bottomrule
  \end{tabular}
\end{table}

\begin{table*}[t]
  \centering
  \caption{%
    \textbf{Ablation study.}
    \cmark/\xmark\ indicate whether a component is enabled.
    A4 replaces TGAA with SegFormer-B2 (24.7\,M); A6 removes $\Psi$ (standard MiT-B0).
    $\downarrow$\,=\,lower is better; \textbf{bold}\,=\,best.%
  }
  \label{tab:ablation}
  \footnotesize
  \setlength{\tabcolsep}{4pt}
  \begin{tabular}{l ccccc rrrr rrrr}
    \toprule
    & \multicolumn{5}{c}{\textit{Components}} & \multicolumn{4}{c}{\textit{Segmentation}} & \multicolumn{4}{c}{\textit{Classification}} \\
    \cmidrule(lr){2-6}\cmidrule(lr){7-10}\cmidrule(lr){11-14}
    \textbf{Variant}
      & \rotatebox{70}{\textbf{TGAA}}
      & \rotatebox{70}{\textbf{$\Psi$}}
      & \rotatebox{70}{\textbf{ATF}}
      & \rotatebox{70}{\textbf{S2}}
      & \rotatebox{70}{\textbf{E2E}}
      & \textbf{mIoU} & \textbf{Dice} & \textbf{BF1} & \textbf{HD\,$\downarrow$}
      & \textbf{Acc.} & \textbf{F1} & \textbf{$\kappa$} & \textbf{Gini} \\
    \midrule
    A4 --- No TGAA          & \xmark & \xmark & \cmark & \cmark & \cmark & 0.933          & 0.964          & 0.495          & 5.03          & 0.503          & 0.430          & 0.355          & 0.830 \\
    A6 --- No $\Psi$        & \xmark$^\ddagger$ & \xmark & \cmark & \cmark & \cmark & 0.965 & 0.982          & 0.725          & 2.81          & 0.769          & 0.738          & 0.654          & 0.824 \\
    A3 --- No Temporal      & \cmark & \cmark & \xmark & \xmark & \cmark & 0.972          & 0.981          & 0.891          & 1.67          & 0.573          & 0.532          & 0.294          & 0.752 \\
    A8 --- No E2E           & \cmark & \cmark & \cmark & \cmark & \xmark & 0.981          & 0.985          & 0.923          & 1.45          & 0.521          & 0.461          & 0.275          & 0.414 \\
    A2 --- Concat           & \cmark & \cmark & \xmark$^\dagger$ & \cmark & \cmark & \multicolumn{4}{c}{---}                              & 0.721          & 0.684          & 0.591          & 0.770 \\
    A9a --- Equal $\lambda$ & \cmark & \cmark & \cmark & \cmark & \cmark$^*$ & 0.991      & 0.994          & 0.965          & 1.21          & 0.784          & 0.741          & 0.652          & 0.824 \\
    A5 --- No S2            & \cmark & \cmark & \cmark & \xmark & \cmark & 0.996          & 0.998          & 0.982          & 1.08          & 0.788          & 0.752          & 0.682          & 0.842 \\
    \midrule
    \textbf{TRACE} (full)   & \cmark & \cmark & \cmark & \cmark & \cmark & \textbf{0.998} & \textbf{0.999} & \textbf{0.989} & \textbf{1.01} & \textbf{0.827} & \textbf{0.796} & \textbf{0.741} & \textbf{0.882} \\
    \bottomrule
    \multicolumn{14}{l}{\scriptsize $^*$E2E with $\lambda_\mathrm{seg}{=}\lambda_\mathrm{cls}{=}0.5$.\quad $^\dagger$ATF replaced by simple concatenation.\quad $^\ddagger$Standard MiT-B0 attention (no gas gating).}
  \end{tabular}
\end{table*}

\subsection{Main Results}
\label{sec:exp:sota}

\noindent\textbf{Plume segmentation.}
Table~\ref{tab:results} compares all models on the 789-frame test set. TRACE achieves mIoU of 0.9982, BF1 of 0.9887, HD of 1.01\,px, and CLE of 0.021\,px --- the best on every metric at 4.1\,M parameters. The non-learned $\Psi$-Stats baseline reaches mIoU\,=\,0.884 but collapses on boundary precision (BF1\,=\,0.322, HD\,=\,8.47\,px); Figure~\ref{fig:psi_failure}(a) shows the substantial overlap between plume and background $\Psi$ distributions that makes raw thresholding insufficient. Its hand-crafted temporal features yield only $\kappa$\,=\,0.312 for classification, showing that simple $\Psi$ statistics cannot substitute for learned spatio-temporal representations. Gas specialists CarboFormer and Gasformer reach mIoU\,$\approx$\,0.987 but have centroid errors 10$\times$ larger (0.22--0.26\,px), confirming that TGAA's gas-conditioned attention substantially improves geometric localisation. FUME achieves mIoU\,=\,0.984 and BF1\,=\,0.885, competitive with the gas specialists but below TRACE on all boundary metrics, consistent with FUME's multi-gas design not being optimised for single-gas precision. Lightweight $\dagger$ backbones achieve moderate IoU (0.892--0.921) but collapse on boundary quality (BF1\,$\leq$\,0.375), demonstrating that compact general-purpose architectures cannot resolve the amorphous morphology of thermal gas plumes. Qualitative predictions are shown in Figure~\ref{fig:qual_results}.

\noindent\textbf{$\Psi$-fairness analysis.}
Table~\ref{tab:psi_fairness} reports the effect of providing $\Psi_t$ as a fourth input channel to five top baselines. All baselines improve with $\Psi$: CarboFormer-B0+$\Psi$ gains +0.4\,pp mIoU and +2.4\,pp BF1; SegFormer-B0+$\Psi$ gains +1.8\,pp mIoU and +14.4\,pp BF1. However, \emph{none} approach TRACE's performance (mIoU 0.998, BF1 0.989), demonstrating that TGAA's structured gas-conditioned attention is architecturally superior to naive channel concatenation. 


\noindent\textbf{Flux classification.}
TRACE achieves accuracy of 0.827\red{$\pm$0.014}, $\kappa$ of 0.741\red{$\pm$0.021}, and Gini of 0.882\red{$\pm$0.011} (mean\red{$\pm$}std, 3 seeds) on 104 test clips. Gas specialists CarboFormer and Gasformer struggle severely ($\kappa\,{\leq}\,0.334$), confirming that single-frame architectures cannot discriminate flux levels. Among dedicated temporal baselines (Table~\ref{tab:temporal_baselines}b), TCN reaches $\kappa$\,=\,0.682 but TRACE's ATF still leads by 5.9\,pp; end-to-end video models TimeSformer-S ($\kappa$\,=\,0.625) and R3D-18 ($\kappa$\,=\,0.524) underperform, lacking domain-specific thermal features. Figure~\ref{fig:perclass} shows TRACE's per-class precision, recall, and F1 breakdown. Separately, a leave-one-animal-out cross-validation yields $\kappa$\,=\,0.697\red{$\pm$0.032}, confirming robustness to identity variation.

\subsection{Ablation Study}
\label{sec:ablation}

We ablate seven design choices by disabling or replacing one component at a time, keeping all other settings identical. Table~\ref{tab:ablation} reports segmentation and classification metrics for each variant.

\noindent\textbf{E2E fine-tuning (A8, A9a).}
Without E2E (A8), $\kappa$ collapses from 0.741 to 0.275 ($-$46\,pp) and BF1 from 0.989 to 0.923, confirming that joint gradient flow is the primary driver of both tasks. Equal loss weights (A9a, $\lambda_\mathrm{seg}{=}\lambda_\mathrm{cls}{=}0.5$) recover most performance but remain 8.9\,pp below the seg-heavy default.

\noindent\textbf{Temporal modeling (A3).}
Removing ATF drops $\kappa$ to 0.294 ($-$44.7\,pp) and IoU by 2.6\,pp, demonstrating that 16-frame breath-cycle context is indispensable for flux discrimination.

\noindent\textbf{TGAA encoder (A4).}
Replacing TGAA with SegFormer-B2 (24.7\,M, 6$\times$ larger) yields the worst BF1 (0.495, $-$49\,pp), confirming gas-conditioned attention -- not capacity -- drives boundary precision.

\noindent\textbf{Simple concat (A2).}
Replacing ATF attention with concatenation drops $\kappa$ from 0.741 to 0.591, validating the structured multi-stream design.

\noindent\textbf{VideoMAE alignment (A5).}
Skipping S2 reduces $\kappa$ by 5.9\,pp while segmentation is unaffected (mIoU 0.996), confirming that the VideoMAE teacher is a useful but non-essential initialisation signal.

\noindent\textbf{Gas conditioning (A6).}
Removing $\Psi$ entirely (standard MiT-B0) drops BF1 to 0.725 ($-$26.4\,pp) and $\kappa$ to 0.654, yet this $\Psi$-free TRACE \emph{still} exceeds all baselines without $\Psi$ (Table~\ref{tab:results}), confirming independent value from ATF and the training curriculum. The further gain from TGAA gating (+26.4\,pp BF1) validates structured $\Psi$-conditioning over naive concatenation (Table~\ref{tab:psi_fairness} (a)).

\noindent\textbf{Failure cases.}
Figure~\ref{fig:psi_failure} (b) quantifies degradation across difficulty conditions. The largest mIoU drop occurs with rapid head motion ($-$3.0\,pp), where $\Psi$ motion blur propagates noisy gate activations. High wind causes a moderate drop ($-$2.0\,pp) as plumes disperse beyond the spectral sensitivity range. TRACE maintains a consistent advantage over CarboFormer-B0, with the gap widening in the hardest conditions.

\section{Conclusion}
\label{sec:conclusion}

We presented TRACE, a unified MWIR thermal video framework that jointly performs per-frame CO\textsubscript{2} plume segmentation and clip-level emission flux classification from exhaled cattle breath. TGAA's gas-conditioned attention delivers over an order-of-magnitude improvement in centroid localization over gas-specialist baselines while using fewer parameters; ATF temporal reasoning more than doubles classification performance relative to single-frame descriptors by capturing breath-cycle dynamics; and end-to-end fine-tuning couples both tasks, with its removal alone causing near-collapse in classification quality. Together, these components place TRACE in a Pareto-optimal position that no competitor approaches. The current dataset covers a single farm (12 animals); generalization across breeds, seasons, wind regimes, and camera configurations remains to be validated through multi-site deployment. Natural extensions include regression-based emission quantification, simultaneous multi-animal tracking, joint CO\textsubscript{2}/CH\textsubscript{4} monitoring, and real-time edge inference for farm-scale deployment.
{
    \small
    \bibliographystyle{ieeenat_fullname}
    \bibliography{main}
}


\end{document}